# Strategies to architect AI Safety: Defense to guard AI from Adversaries

Rajagopal. A[1], Nirmala. V[2]

*Abstract*

The impact of designing for security of AI is critical for humanity in the AI era. With humans increasingly becoming dependent of AI, there is a need for neural networks that work reliably, inspite of Adversarial attacks. This paper aims to build defenses for 3 types of attacks:

1. Similar looking adversarial images that aim to deceive both human and computer intelligence,
2. Adversarial attacks such as evasion and exploratory attacks,
3. Hacker introduced occlusions or perturbations to misguide AI

The vision for Safe and secure AI for popular use is achievable. To achieve safety of AI, this paper explores strategies and a novel deep learning architecture. To guard AI from adversaries, paper explores combination of 3 strategies:-

1. Introduce randomness at inference time to hide the representation learning from adversaries.
2. Detect presence of adversaries by analyzing the sequence of inferences.
3. Exploit visual similarity against adversarial perturbations.

To realize these strategies, a novel architecture is designed. This Dynamic Neural Defense, DND has 3 deep learning features:

1. By hiding the way a neural network learns from attacks using a random computation graph, DND evades attack.
2. By analyzing input sequence to cloud AI inference engine with LSTM, DND detects attack sequence.
3. By inferring with visual similar inputs generated by VAE, it does not succumb to hackers.

Thus, a roadmap to develop reliable, safe and secure AI is presented.

## I. INTRODUCTION: WHY AI SAFETY

### A. The importance of security of AI

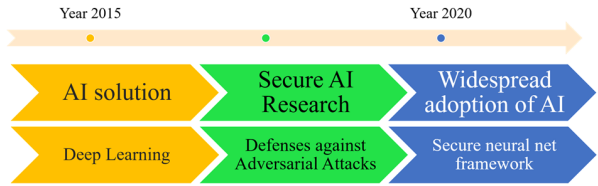

Figure 1. Why safe AI is critical for adoption of AI

The field of Artificial Intelligence is witnessing amazing progress thanks to extraordinary work in research in academia and open source Deep Learning frameworks. But the practical widespread adoption of Deep Learning based solutions is still yet to happen. One of the key challenges is that they are vulnerable to Adversarial Attacks [1]. Research on AI security will open doors for widespread adoption of AI as per Fig 1.

### B. Purpose of the Contribution: Security of AI

The goal of this paper is to improve safety of AI by proposing a novel approach that learns to become resilient inspite of:

1. Adversarial attacks such as evasion and exploratory attacks on CNNs;
2. Adversarial attacks that introduce perturbations to create visually similar looking adversarial images that aim to deceive both human and AI; and
3. Hacker introduced occlusions or alterations in physical environment to misguide AI decisions.

### C. Is it safe to depend upon AI?

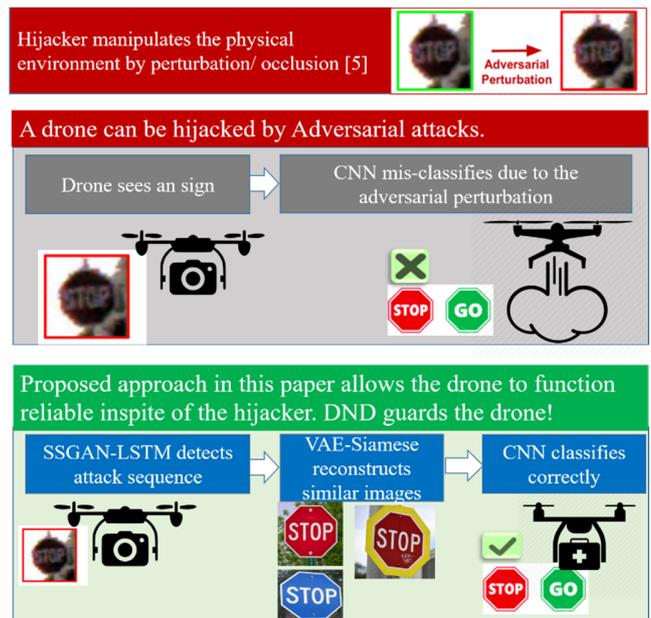

Figure 2. With and without Safe AI.

With and without safe AI, the outcome can be totally different as illustrated in Fig 2. Is it safe to depend upon AI?

1. Can an autonomous self-flying drone be hijacked by rival introduced perturbations?
2. Can a drone take down a terrorist walking on the road?

The enormous potential of Deep Learning can be realized for masses only when it is depend-able. But researchers are beginning to ask the question: Is it depend-able? Can mankind depend on decisions made by AI. When mankind begins to depend upon a new technology, it shouldn't be vulnerable. A reliable technology needs to work irrespective of the circumstances.

### D. Contribution of the paper

Towards achieve the goal of guarding AI, this paper discusses on two aspects
1) Strategies to design safety into AI; and
2) Novel neural network architectural framework to realize the strategies.

The above two are summarized in Fig 3. The proposed strategy presented in this paper offers a significant potential to improve resilience of AI systems against adversaries, setting a forward vision for safe AI in mainstream deployments. In addition, the paper also explores a novel approach in architecting neural network to realize the strategies.

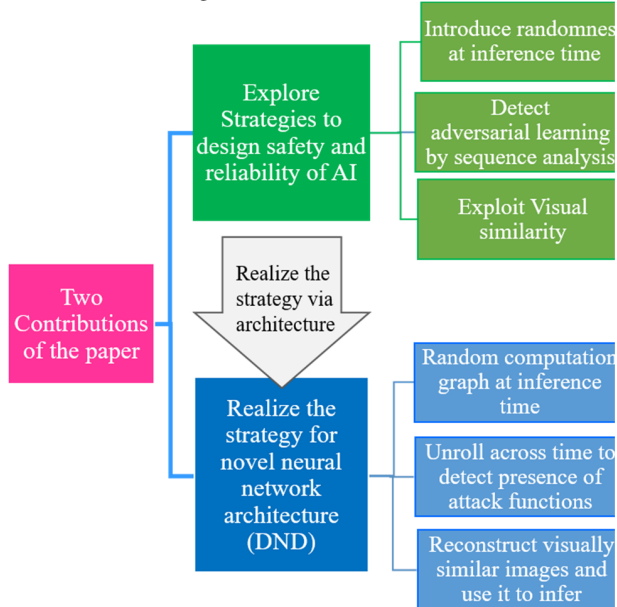

Figure 3. Strategy and AI Architecture

## II. STRATEGIES FOR AI SAFETY

This paper proposes strategies to defend AI against adversaries. To explore the insights, Table 2 discusses it.

The 3 strategies are

1. Introduce randomness to hide the learning function from adversaries:

    a. Through a random computation graph at inference time, this defense strategy introduces randomness to hide the learning function from an attacker.
    b. Neural architecture search automatically identifies a set of network graphs that optimizes for minimizing transferability of adversarial examples.

2. Detect adversaries by analyzing the sequence of inputs:

    a. As adversaries can attack only after a sequence of inference requests to the cloud served model, the defense strategy is to analyze by unrolling across time to detect presence of attack functions.
    b. Further, a decoy neural network is used to early detect such adversaries

3. Exploit visual similarity against adversarial perturbations:

    a. Like humans, this approach looks for visual similarity to improve robustness inspite of adversarial perturbations. It reconstructs visually similar images and infers from a random reconstruction, thus insulating the AI for any attacking systems that may try to learn about the model. This idea is used in DND of Fig 5.
    b. In addition, this mechanism improves robustness as inference is based on a visually similar images rather than the adversarial input.

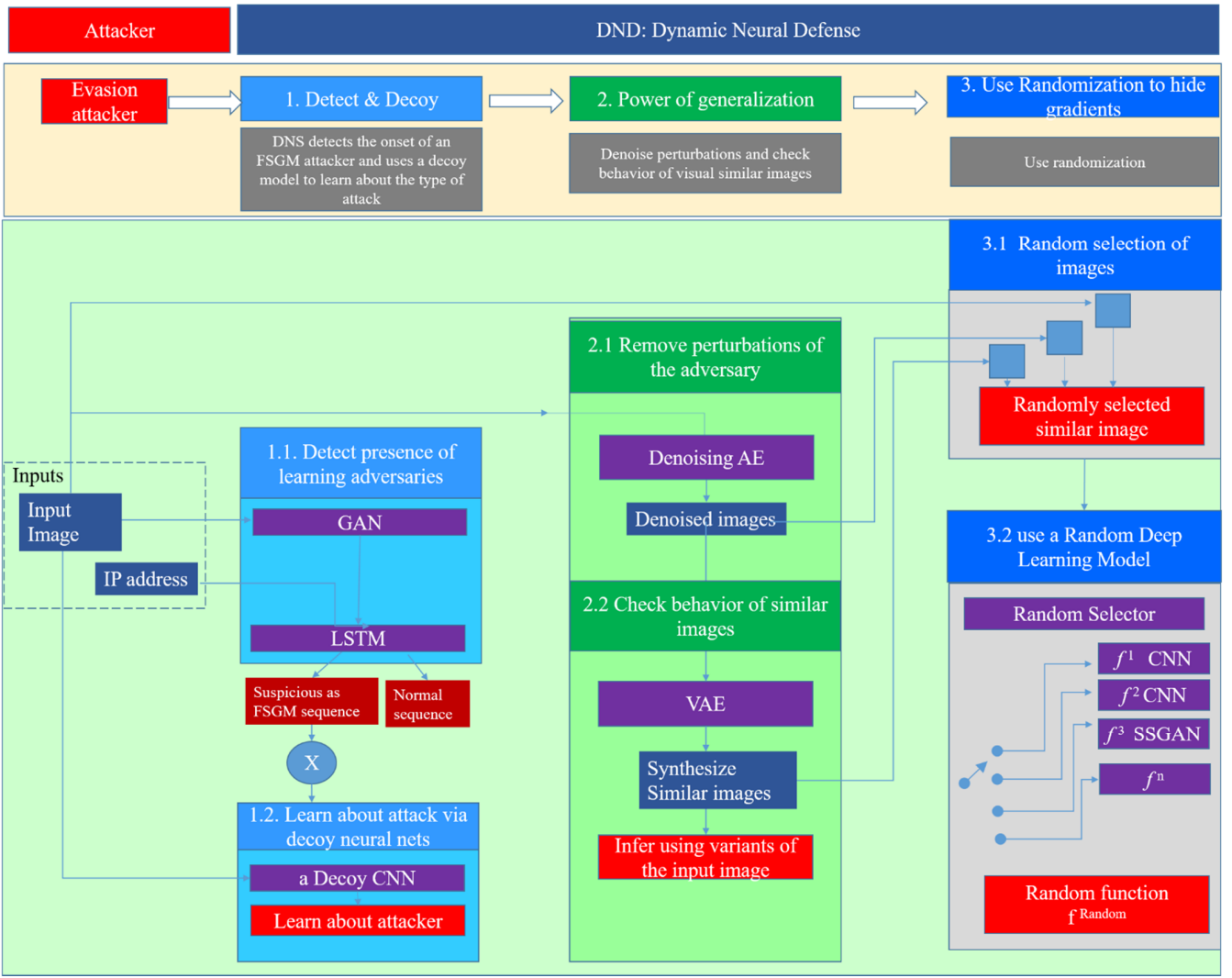

Figure 4. Proposed DND architecture to enable Safe AI.

### III. NEURAL DEFENSE ARCHITECTURE FOR SAFE AI

To realize this strategy, the paper also presents a novel architectural approach as illustrated in Fig 4. The strategies are realized as a novel architecture, called Dynamic Neural Defence (DND). DND, a Neural network approach has three major deep learning based architectural features:

1. By hiding the way a neural network learns from attacks, DND evades attack. DND dynamically infer by a random choice of neural network (or a randomly selected ensemble) at run-time or a dynamically choosen computation graph. This stochastic computation hides the learning algorithm from an attacker. Thus, a three layers of "**AI firewall**" is illustrated in Fig 5. One may add 4[th] layer with SHIELD[16].

 A Neural Architecture Search is employed to design a set of CNN models in such a way, so that transferability of adversarial examples is minimized. Refer module 3.2 of DND architecture presented in Fig 4.

2. DND can learn to adapt dynamically by learning about occlusion/attack patterns from a sequence of attacks via SSGAN-LSTM detector that detects the presence of a fast gradient sign attack sequence. Refer block 1.1 in Fig 4.

3. DND not just only denoises using AE, but also infers using additional set of visual similar inputs generated by VAE with small variance, thus improving reliability of classification by inferring through other examples that are visual similar to the adversarial input. Please refer block 2.1 and 2.2 in the architecture presented in Fig 4.

Table 1: Nomenclature used in this paper

| | |
|---|---|
| DNN | Deep neural networks |
| *Safe*AI | Safety of AI systems also Robustness of DNN |
| DND | A proposed architecture in this paper called Dynamic Neural Defense |
| $f$ | Deep Learning Model |
| **(X,y)** | Input pair in the dataset of unsupervised Deep Learning |
| **W** ← train($f$, X,y) | Set of Weights matrix of DNN learnt after training the model $f$ on **(X,y)**. |
| **y**$^{pred}$ = $f$ (**W*X**) | Predicted label when a Deep Learning Model, $f$ infers a input,X using the learnt Weights,**w** |
| NAS | Neural Architecture Search Algorithm |
| $f^1$, $f^2$ | Two different Deep Learning Models created by NAS |
| $f^n$ | A n$^{th}$ Model $f$ created by NAS |
| *Attacker* | Adversarial Attacks on Deep Neural Networks [1] |
| *Defender* | Defender with a goal of *Safe*AI and applies Defences against *Attacker* [1] |
| $f^{[Random\ i]}$ | A *Defender* selects a random i$^{th}$ Model $f^i$ at inference time and uses it to classify |
| ATM | Adversarial Threat Model [1] |
| WBA | White Box Attacks [1] |
| FGSM | Fast Gradient Sign Method based Attack [3] |
| *Attacked $f^1$* | *Attacker* learns about the vulnerability/gradients of 1$^{st}$ Model $f^1$ about using FSGM |
| CNN | Convolutional Neural Network |
| LSTM | Long Short Term Memory Recurrent Neural Network |
| AE | AutoEncoder |
| VAE | Variational AutoEncoder |
| SSGAN | Semi Supervised Learning using Generative Adversarial Network |

Table 2: Summary of the key ideas in this paper

| Insights on how to defend | idea | Proposed Novel approach to Defense |
|---|---|---|
| 1. Make it difficult for Evasion Attacker to gain knowledge about how DNN behaves | **Introduce randomness at inference time** | The inference engine uses a random Deep Learning Model $f^{[Random\ i]}$ (**W**) for each inference, thus masking the gradients from attackers. 3 layers of **AI firewalls** block flow of gradient information to any attack, as illustrated in Fig 5. |
| 2. Remove the perturbations of the Adversary | **DNN learns to a remove perturbations** | De-noising Autoencoders remove perturbations/noise as shown in block 2 in Fig 4. Presence of other attacker's objects in the scene are identified with YOLO, and hence deliberately introduced occlusions are detected & even removed, improving resilience of AI. |
| 3. Predict using a visually similar images such as an input | **Generate visually similar images & classify using them** | Inference engine employs Variational autoencoders with minor variance to generate more similar looking variants of the input, and then infers from them. With varience, the DNS shields the gradient from the adversary as Fig 5. |
| 4. Detect presence of Attacker by analysing sequence of inference | **Detect presence of functions such as FGSM** | Watch a sequence of images from an IP address, and detect the presence of FGSM algorithms. If so, deploy a decoy to learn about the attack algorithm as shown in block 1 in Fig 4. |
| 5. Block transferability of adversarial examples by learning to identify them | **Generate Adversarial examples & learn to detect them ahead of deployment.** | DND can anticipate how a typically classifier can be evaded using adversarial examples, and uses this understanding to avoid taking actions under such adversarial situations. An over fitted classifier which has memorized to detect adversarial examples, along with a semi supervised GAN to senses adversarial examples. |

## IV. RESULTS

To enable safety & security of AI, this paper's discussed strategic ideas and novel architecture framework. The key ideas are summarized in Fig 3.

A path for secure AI is feasible.
1) Insights to enable AI Safety are discussed in Table 2.
2) Novel Deep Learning architecture to enable AI Safety is conceptualized & visualized in Fig 4.

## V. CONCLUSION AND FUTURE SCOPE

### A. Why contribute to Safety of AI ?

The imperative to design for security in early stages of deep neural networks and not an after-thought especially with the risk of hacker exploiting powerful autonomous AI. The black box nature of neural networks make it difficult to even detect that a neural network has been hacked into, and even pose a great danger for human exist

### B. Contributions: Roadmap, Strategies & Neural Net

This paper offers a potential to improve power of resilience of AI systems against adversaries, setting a forward roadmap for safe usage of AI by one and all. Together with insights, ideas and neural architecture principles, the roadmap to architect for reliable AI is offered by this paper.

Summary of key contributions
1) The paper contributions are summarized in Fig 3.
2) A roadmap for Safe AI is presented in Table 2
3) Neural defence is conceptualized in Fig 4 & Fig 5.

### C. Future directions: Defense of the future

Safety of AI is an emerging topic for AI researchers. Research on ideas discussed in Table 2 can lead to breakthrough

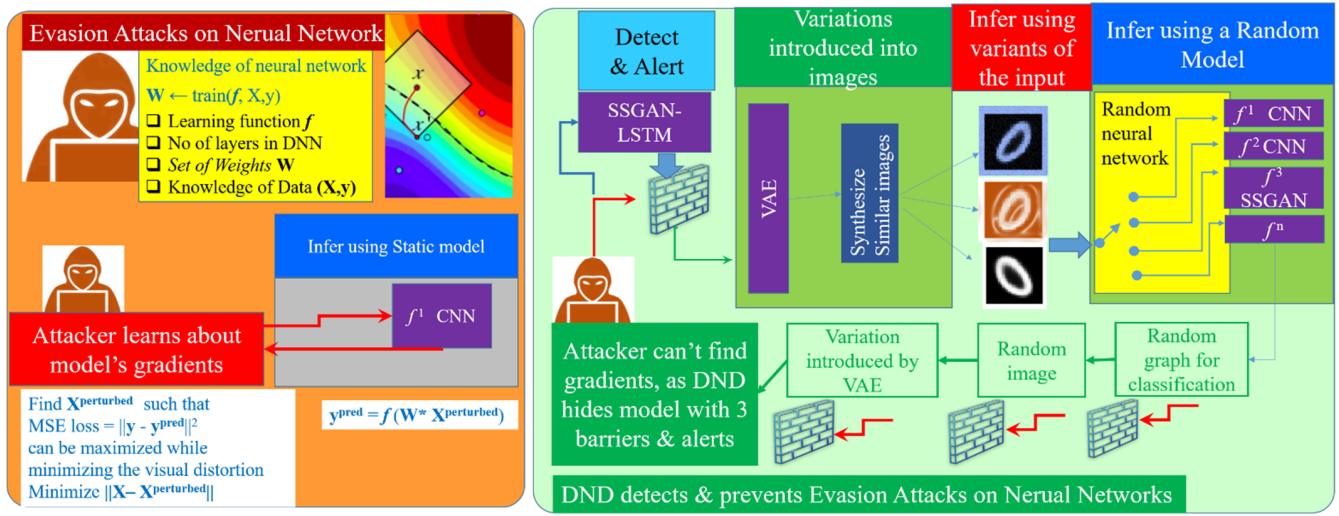

Figure 5. How the proposed architecture hides from an attacker?

## Comments

1 rajagopal.motivate@gmail.com
Indian Institute of Technology, Madras
2 gvan.nirmala@gmail.com (correspondance)
Queen Mary's College